\documentclass{article}

\usepackage{arxiv}

\usepackage[utf8]{inputenc} 
\usepackage[T1]{fontenc}    
\usepackage{hyperref}       
\usepackage{url}            
\usepackage{booktabs}       
\usepackage{amsfonts}       
\usepackage{nicefrac}       
\usepackage{microtype}      
\usepackage{graphicx}
\usepackage{natbib}
\usepackage{doi}

\title{Recurrent and Convolutional Neural Networks in Classification of EEG Signal for Guided Imagery and Mental Workload Detection}

\author{ Filip Postepski\thanks{Maria Curie-Sklodowska University, Department of Neuroinformatics and Biomedical Engineering, Institute of Computer Science, Akademicka, Lublin, 20-031, Poland}\\
	Maria Curie-Sklodowska University\\
	\texttt{filip.postepski@mail.umcs.pl} \\
	\And
	Grzegorz M. Wojcik*\\
    Maria Curie-Sklodowska University\\
	\texttt{gmwojcik@live.umcs.edu.pl} \\
	\And
	Krzysztof Wrobel*\\
    Maria Curie-Sklodowska University\\
	\texttt{krzysztof.wrobel@mail.umcs.pl} \\
	\And
	Andrzej Kawiak*\\
    Maria Curie-Sklodowska University\\
	\texttt{andrzej.kawiak@mail.umcs.pl} \\
	\And
	Katarzyna Zemla\thanks{SWPS University, Institute of Psychology, Chodakowska, Warsaw, 03-815, Poland}\\
    SWPS University\\
	\texttt{kzemla1@st.swps.edu.pl} \\
    \And
    Grzegorz Sedek\textdagger\\
    SWPS University\\
	\texttt{gsedek@swps.edu.pl} \\
}

\renewcommand{\shorttitle}{RNNs and CNNs in Classification of EEG Signal...}

\hypersetup{
pdftitle={Recurrent and Convolutional Neural Networks in Classification of EEG Signal for Guided Imagery and Mental Workload Detection},
pdfsubject={cs},
pdfauthor={Filip Postepski, Grzegorz M. Wojcik, Krzysztof Wrobel, Andrzej Kawiak, Katarzyna Zemla, Grzegorz Sedek},
pdfkeywords={guided imagery , mental workload, EEG, CNN, LSTM},
}

\begin{document}
\maketitle

\begin{abstract}
	The Guided Imagery technique is reported to be used by therapists all over the world in order to increase the comfort of patients suffering from a variety of disorders from mental to oncology ones and proved to be successful in numerous of ways. Possible support for the therapists can be estimation of the time at which subject goes into deep relaxation. 
This paper presents the results of the investigations of a cohort of 26 students exposed to
Guided Imagery relaxation technique and mental task workloads conducted with the use of
dense array electroencephalographic amplifier. The research reported herein aimed at verification whether it is possible to detect differences between those two states and to classify them using deep learning methods and recurrent neural networks such as EEGNet, Long Short-Term Memory-based classifier, 1D Convolutional Neural Network and hybrid model of 1D Convolutional Neural Network and Long Short-Term Memory. 
The data processing pipeline was presented from the data acquisition, through the initial data cleaning, preprocessing and postprocessing. The classification was based on two datasets: one of them using 26 so-called cognitive electrodes and the other one using signal collected from 256 channels. So far there have not been such comparisons in the application being discussed. 
The classification results are presented by the validation metrics such as: accuracy, recall, precision, F1-score and loss for each case. It turned out that it is not necessary to collect signals from all electrodes as classification of the cognitive ones gives the results similar to those obtained for the full signal and extending input to 256 channels does not add much value. 
In Disscussion there were proposed an optimal classifier as well as some suggestions concerning the prospective development of the project.
\end{abstract}


\section{Introduction}
Relaxation methods proved to be helpful for the patients with some
illnesses and mental disorders. Oncological patients were reported to respond better to treatment when they used relaxation techniques\cite{breastCancer}. Therefore, it is
beneficial to develop relaxation techniques in order to improve the quality of life. Moreover Guided Imagery can be used as relaxation technique. It is largely applied and proved to be effective in reducing test anxiety and dealing with stress of different origins\cite{relaxPreg, stephens1992imagery, nguyen2018nature}.  
Electroencephalography (EEG) can be a good method to find out if patients are in the state
of relaxation or not. Scalp EEG is a non-invasive method of measuring bio-electrical activity of the
human brain. Moreover, it is less expensive and less stressful for patients than other
brain activity measuring devices, such as PET or MRI\cite{MURPHY199751,
eeg_sig_proc_and_ml}. On the other hand, manual multichannel EEG signal analysis can be a
difficult and time-consuming process. Machine learning and deep learning tools are
commonly used to classify various types of data, starting with the images\cite{imageNet} to the different kinds of
signals\cite{rcnnClass, cheng_ecg_2021}. The aim of this study is to
propose an EEG signal classifier based on the 1D Convolutional Neural Networks (CNNs) by using raw signal with only basic filtering done as an input data.

For different types of EEG signals, classical machine learning (ML) methods, such as Support
Vector Machines (SVM), were used \cite{svmEEG}. In classification of relaxation and concentration states based on the electroencephalographic signal SVMs can achieve around 80\% of accuracy (ACC)\cite{svmRelaxation}. 

State-of-the-art classification methods applied for the EEG signal already used Convolutional Neural Networks (CNNs) with success \cite{cnnAppOh}. Furthermore, the above mentioned classical ML methods are increasingly being replaced by deep learning approaches. Convolutional Neural Networks are applicable in the EEG signal analysis, for instance, in motor imagery processing \cite{xu2019deep}, epileptic seizure detection \cite{zhou2018epileptic}, emotion recognition \cite{zhang2020investigation}, and research topics devoted to Brain-Computer Interfaces based on EEG feature extraction using CNNs \cite{chen2023brain}, among others, even for identity authentication \cite{zhang2022eeg}. 

The most common approach is to classify signals by feeding the classifier
with the frequency bands data. The EEG signal is commonly partitioned into discrete frequency ranges, encompassing delta waves below 4 Hz, theta waves ranging from 4 to 7 Hz, alpha waves spanning 8 to 12 Hz, beta waves between 13 and 30 Hz, and gamma waves surpassing 30 Hz.  It was proved that using specific selected bands of EEG signal, SVM classificator can be done\cite{svmRelaxation, shengEEGclassCNNfeature}. Calculation of power across specific frequency bands is needed. Therefore it would be beneficial to skip manual feature extraction and use CNN-based feature extraction from the raw signal. Some researchers used this approach successfully for emotions recognition\cite{chenRawEEG,YanagimotoRawEEG}. The experiments described by Baydemir et al. showed that it is possible to classify EEG signal of low and high cognitive load using 1D-CNN with a great accuracy\cite{baydemir2022classification}. Classification of fNIRS-EEG mental workload signal using CNN was made, showing a good accuracy of 89\%\cite{fnirs_eeg}. However there are only few papers including 1D Convolutional Neural Networks used specifically in the binary classification of relaxation and mental workload using the raw EEG signal which still needs to be investigated.

In our previous research, the classical classification method was used for Guided Imagery and Mental Task groups \cite{zemla2023-glm}. Generalized Linear Model (GLM) used in that research achieved 81\% accuracy using a very specific time segment, 779-839 seconds, extracted from the complete recording. In order to achieve this level of accuracy, this required feeding the classifier with five EEG bands (alpha, beta, delta, theta, and gamma), extracted from the raw signal of the 60 seconds duration. However, on the full-length recording, the accuracy of 90.77\% was achieved.

The objective of this study is to compare four approaches to classification of EEG signals of two mental states: Guided Imagery relaxation technique and Mental Workload tasks. For this research 1D Convolutional Neural Network (1D-CNN), Long-Short Time Memory (LSTM), 1D-CNN-LSTM hybrid model and 2D-CNN (EEGNet) will be taken into consideration. Signals were filtered and split into 1-second segments. Bad channels were marked automatically and interpolated. That way all 256 channels could have been used for training. No further preprocessing or artifact removal was done. No features were extracted from that signal manually. 

\section{Materials and methods}
The signal for this study was obtained from a cohort of 26 males, aged 19-24 years. They
were all right-handed and short-haired. Being right- or left-handed could influence the
results due to brain lateralization. Described experiments were reviewed and approved by the Maria Curie-
Skłodowska University Bioethical Commission. The experiments were conducted according to the best experimental practices and guidelines. They were also done under the supervision of qualified psychologists. All participants agreed to the EEG signal recording and were
informed about the purpose of the experiment. They all signed written consent before taking part in it. 

\subsection{Inclusion and exclusion criteria}
The criteria for selecting participants in this study involve being a healthy, right-handed male, aged 19 to 24, with short hair and fluency in Polish. They should have no history of chronic diseases, no current use of prescribed or recreational drugs, and should be able to attend study appointments without specific technological requirements. Additionally, participants were required to abstain from alcohol and medication for at least 72 hours before the experiment.

On the other hand, exclusion criteria encompassed individuals younger than 19 or older than 24, left-handed individuals, those with long hair, limited proficiency in Polish, serious or chronic illnesses, current use of medications or drugs, recent medical treatments, or inability to attend study appointments. Participants failing to meet the inclusion criteria or declaring serious diseases, including mental disorders, were automatically excluded. Prior to participation, participants were informed about the EEG research and technology and consented to take part in the study.

There were several reasons for recruiting participants aged 19-24 and only males. Firstly, the majority of individuals in this age range are students, particularly those pursuing first and second degrees. Secondly, in the Institute of Computer Science, there is a predominant male student population, making it challenging to form both target and control groups including women. However, the most significant reason was the documented changes in women's EEG cortical activity throughout the menstrual cycle, as published by \cite{solis1994eeg, krug1999variations}. These changes introduce additional variables into the model. Variations are observed in both alpha and beta bands \cite{bazanova2014eeg, souza2022effect}, which could be crucial for signal classification related to the individual's state of mind.

Moreover, it was noted that a substantial majority of female computer science students had lengthy hair. It is noteworthy that the research has also highlighted differences in electroencephalogram patterns between males and females \cite{wada1994gender, cantillo2017gender}, and the objective was to achieve a relatively balanced representation from the participant pool.

They all signed a written consent. Half of the group listened to the Guided Imagery relaxation recording prepared by the psychologist. The other half were asked to recall specific kinds of information: the
names of Polish administrative units (voivodships), the names of the Zodiac signs, the names
of US states, etc. (Mental Task group or MT group). Tasks were given by the same psychologist on the recording. After each task there was a period of silence when participants were thinking about the answer. The GI group was supposed to relax during
the experiment, while the MT group was supposed to be put under mental workload. At the
beginning of the experiment the MT group was told that after its completion they would be asked to write down all
the information they will have recalled. The Guided Imagery and the Mental Task recordings
were of the same length of 20 min. The participants were asked to close their eyes and each
trial was conducted in the lying position with lights turned off to decrease the effects of
muscle artifacts, power line noise and distractions on the EEG signal.

\begin{figure}
\centering
        \includegraphics[width=0.9\linewidth]{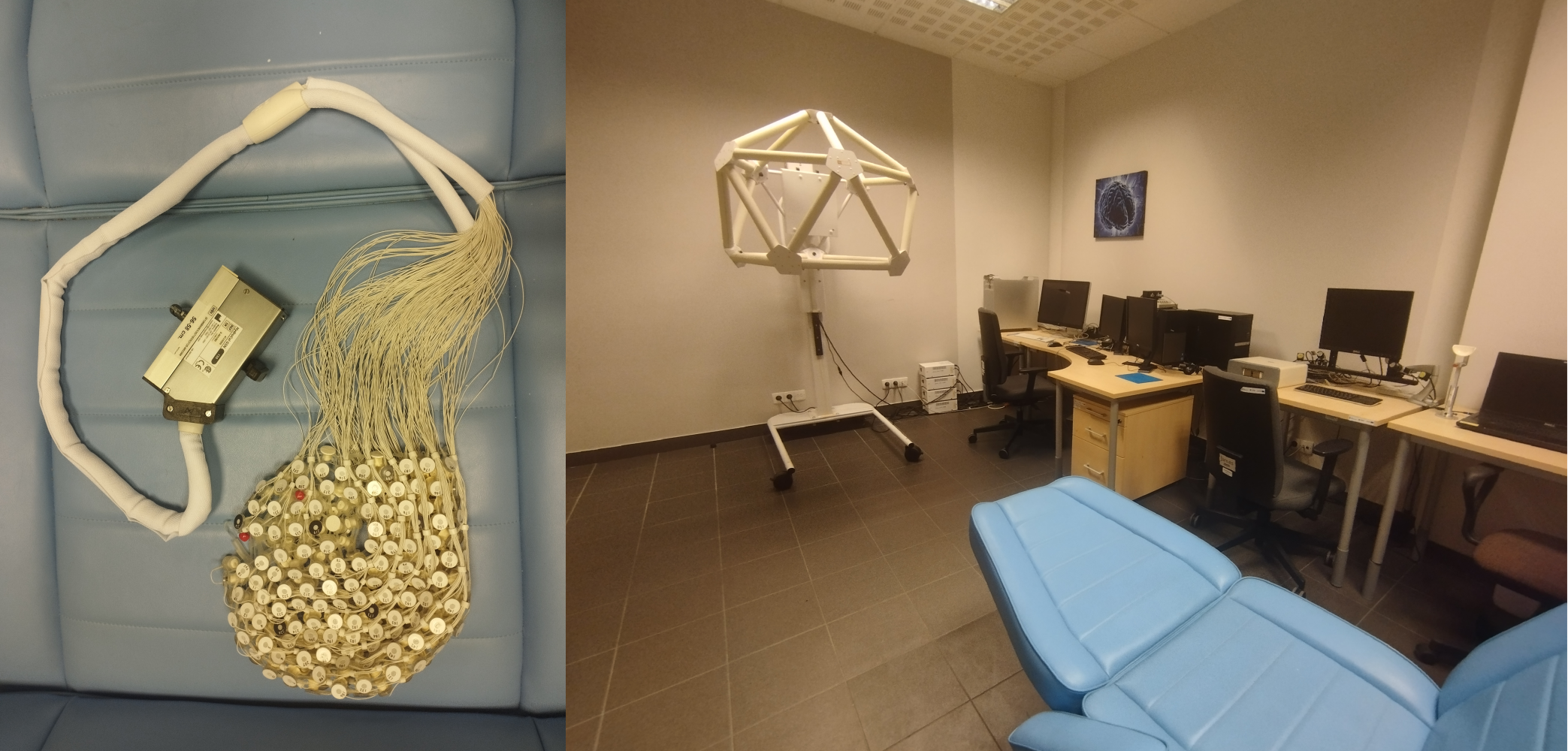}
        \caption{On the left: EGI 256-channel EEG cap.On the right: the overview of the whole EEG Laboratory at UMCS, Lublin, Poland}
        \label{lab}
\end{figure}

The experiments were conducted in the EEG Laboratory of the Department of
Neuroinformatics and Biomedical Engineering of Maria Curie-Skłodowska University (UMCS)
in Lublin, Poland (Figure \ref{lab}). All trial signals were recorded at the sampling frequency
of 250 Hz with the use of a 256-channel dense array EGI GSN 130 series cap (Figure \ref{lab}). For
signal acquisition, the EGI Net Station 4.5.4 software was used.

\begin{figure}
    \centering
    \includegraphics[width=0.8\textwidth]{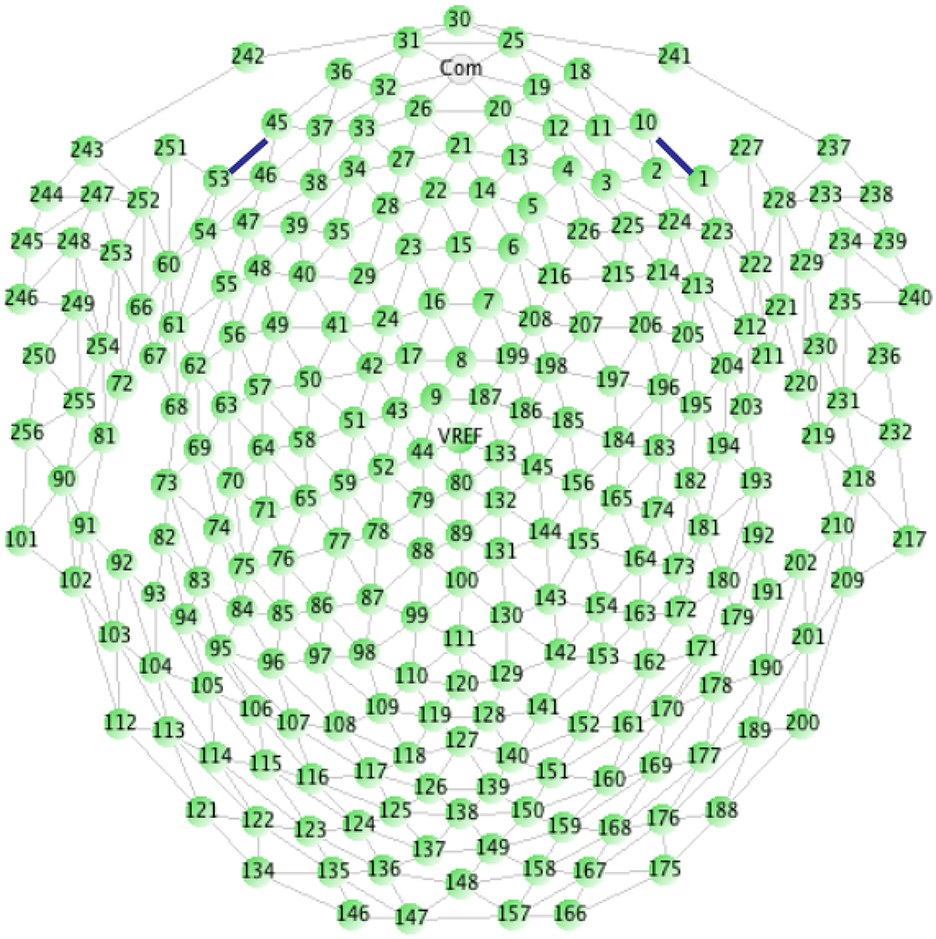}
    \caption{Electrodes placement on HydroCel GSN 130 Geodesic Sensor Net\cite{geodesics2009geodesic, wojcik2023investigating}}
    \label{fig:scalp-egi}
\end{figure}

Our dense array amplifier recorded the signal from all 256 electrodes. However, we expected to find differences on the so-called cognitive electrodes based on the previous experience in the cognitive processing EEG signal analysis \cite{wojcik2023investigating, kawiak2020whom, kwasniewicz2021believe, schneider2022modeling}. These electrodes are described in the EGI 256-channel cap specification \cite{geodesics2003net, geodesics2009geodesic, geodesics2011geosource} as the best for cognitive ERP observations, covering the scalp regularly, and numbered as follows: E98, E99, E100, E101, E108, E109, E110, E116, E117, E118, E119, E124, E125, E126, E127, E128, E129, E137, E138, E139, E140, E141, E149, E150, E151, and E152 (see Fig.\ref{fig:scalp-egi}).

\subsection{Signal preprocessing and data sets preparation}

\begin{figure}
    \centering
    \includegraphics[width=0.8\linewidth]{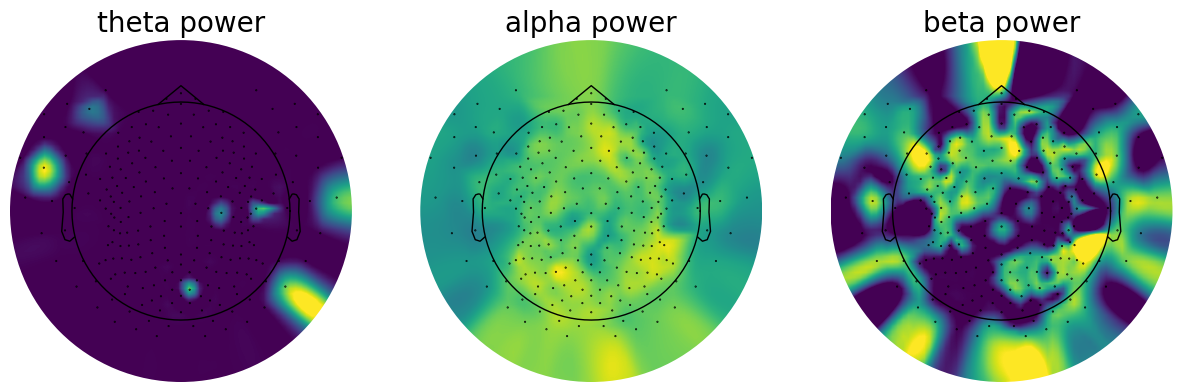}
    \caption{Power spectral density of different frequency bands shown for 1-s segment of signal from GI sample subject}
    \label{power_spectral1}
\end{figure}

\begin{figure}
    \centering
    \includegraphics[width=0.8\linewidth]{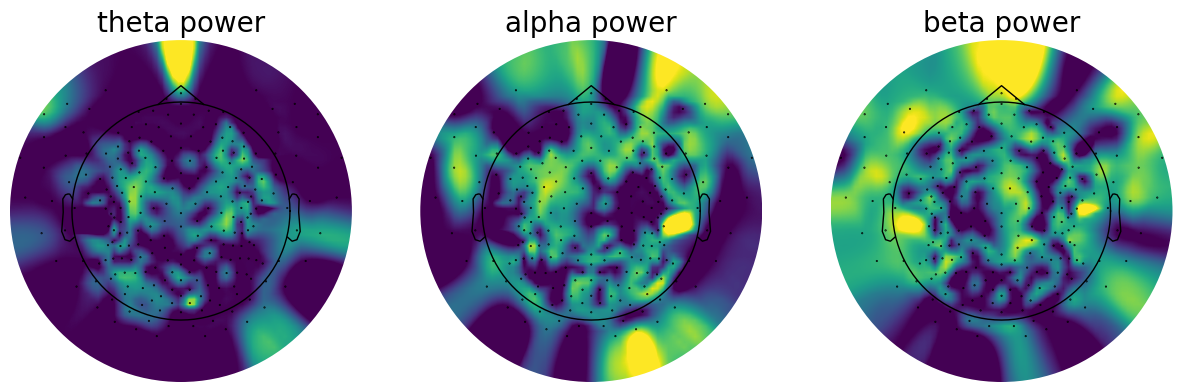}
    \caption{Power spectral density of different frequency bands shown for 1-s segment of signal from MT sample subject}
    \label{powers_spectral2}
\end{figure}

\begin{figure}
    \centering
    \includegraphics[width=0.8\linewidth]{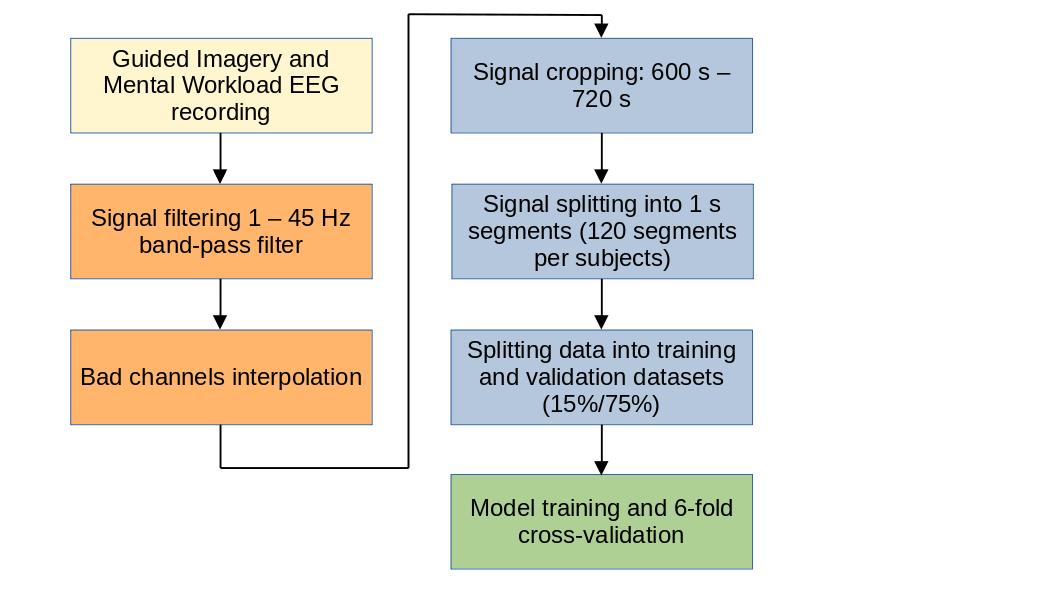}
    \caption{Data science pipeline - steps of data preparation for training}
    \label{data_science_pipeline}
\end{figure}

The recorded EEG signals were pre-processed using mne Python toolkit 1.3.0
\cite{mnePython}. Noisy channels were removed from the signal and interpolated to
maintain the same size of data in each sample. For automatic bad channel rejection the RANSAC algorithm implemented in pyprep toolkit \cite{appelhoff_2023_10047462pyprep} was used. This toolkit is based on the PREP pipeline designed for EEG signal preprocessing in MATLAB\cite{fninf.2015.00016prep}. The signal from each trial was filtered with a
band pass filter of 1-45 Hz. Each signal was cropped from 10 to 12 minutes of the
recording, which gives 120 seconds per subject. The time segment was chosen based on the previous experience with GI relaxation method. It was proved that the period between 10 and 14 min. of recording has the greatest significance for distinguishing the relaxation and mental workload state \cite{zemla2023-glm}. Each cropped signal was split into 1-second segments. This gives a total amount of 3,120 recording samples (1,560 samples of Guided
Imagery group and 1,560 samples of Mental Task group). Figure \ref{data_science_pipeline} shows the data preparation steps. The sample 1-s segments for both GI and MT states were shown in terms of different power densities for each of frequency bands in Figures ~\ref{power_spectral1} (for GI) and ~\ref{powers_spectral2} (for MT).

Two sets of electrodes were selected for the experiments. The first one included a full set of 256 channels of EEG signal. The second one contained a subset of 26 electrodes from the central-parietal region to reduce the amount of data subjected to training. Based on the previous research in analyzing cognitive processing of EEG signals
\cite{kawiak2020look, kwasniewicz2020you, schneider2022modeling}, variations were
expected to be observed specifically on the above mentioned 26 cognitive electrodes.  Those electrodes, specified as optimal for observing cognitive phenomena according to the EGI 256-channel cap specifications \cite{waveform}, are positioned in the central-ocipital region and numbered: E98, E99, E100, E101, E108, E109, E110, E116, E117, E118,
E119, E124, E125, E126, E127, E128, E129, E137, E138, E139, E140, E141, E149, E150, E151,
and E152. The topographical map showing the placement of these electrodes on the scalp
can be found in the EGI documentation \cite{waveform} and in
\cite{wojcik2023investigating}, Fig. ~1. It was also showed that they cover the region of the greatest significance for the alpha band-based research, as this band is correlated with the relaxation state\cite{eeg_sig_proc_and_ml}.
Finally, the both datasets consisted of 3,120 signal samples. Each sample included 256 EEG channels in the data set 1 (FULL-256) or 26 EEG channels in the data set 2 (COGN-26), and 250 timesteps per second. No further pre-processing or feature extraction was done.

The data set was split into 2,640 samples in the training
data set and 480 samples in the testing data set. 6-fold cross-validation was used to confirm performance of the model. The StratifiedGroupKfold method from scikit-
learn\cite{scikit-learn} was used to prevent the data from one subject to be put in training and validation data sets at the same time. On the other hand, StratifiedGroupKFold keeps the data set with a balanced number of samples for each group. The data set was shuffled to prevent the model
from learning data from only one subject in one batch. Folds were saved for benchmarking purposes. 

\subsection{EEGNet}
\begin{figure}
    \centering
    \includegraphics[width=1\linewidth]{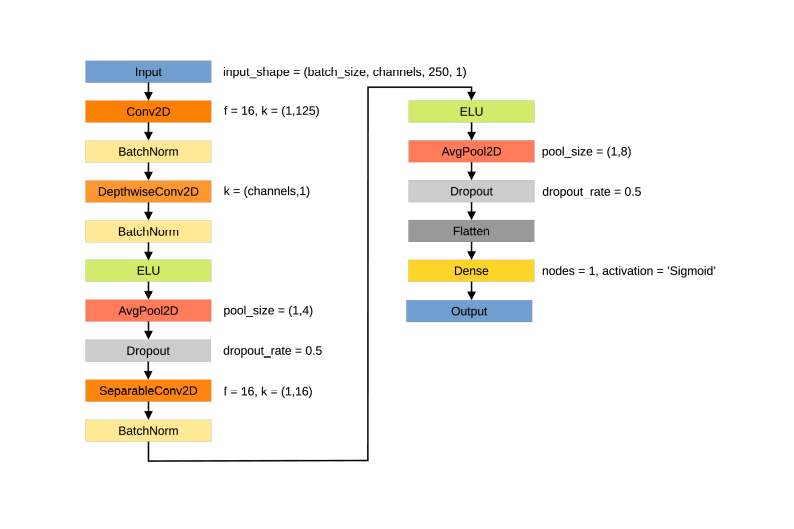}
    \caption{EEGNet detailed model architecture with parameters for specific layers: f - number of filters, k - kernel size, pool sizes and dropout rates. }
    \label{fig:eegnet_model_architecture}
\end{figure}

The first method of classification of EEG signal in this research was 2D-CNN architecture called EEGNet proposed by Lawhern et al.\cite{Lawhern_2018_eegnet}. Implementation of this network was done using tensorflow and keras. All architecture remained as presented in the original research. The parameters were adjusted as suggested by the EEGNet authors. All parameters are described in table Table ~\ref{tab:table_eegnet} and are given in Figure ~\ref{fig:eegnet_model_architecture}.

The learning rate was set to 0.001, the optimizer was Adam and the loss function was binary cross-entropy. Loss function selection resulted in changing the activation function from original Softmax to Sigmoid. 

EEGNet performance in terms of validation accuracy and validation loss was selected as reference for all other methods of binary classification described in this research. Using COGN-26 data set, the model had 2,153 parameters. After training on FULL-256 data set the model had 6,753 parameters.

\begin{table}[!ht]
    \centering
    \begin{tabular}{ |l|l|l| }
        \hline
        \textbf{Parameter}& \textbf{Description} & \textbf{Value}\\ 
        \hline
        F1& Number of temporal filters & 8\\
        \hline
        F2& Number of pointwise filters & 16\\
        \hline
        k& Kernel length & 125\\
        \hline
        D& Number of spatial filters for each temporal convolution & 2 (original value)\\
        \hline
        -& Activation function in output layer& Sigmoid\\
        \hline
    \end{tabular}
    \caption{Parameters set for EEGNet architecture according to original paper\cite{Lawhern_2018_eegnet}}
    \label{tab:table_eegnet}
\end{table}

\subsection{LSTM}

Long short-term memory (LSTM) is a type of Recurrent Neural Network cell introduced as a solution for learning features from long time sequences including noisy data\cite{hochreiter1997long}.

The simple LSTM-based network was tested as a second reference method. It was proved that Bidirectional LSTM-based (BiLSTM) model can be a good method of EEG classification tasks like emotion classification\cite{YANG2020491emotion_lstm} or seizure classification\cite{hu_scalp_2020_lstm}.

The architecture presented here contained one BiLSTM layer having 64 units(cells) for each backward and forward direction. The number of units were selected according to \cite{YANG2020491emotion_lstm}. As the input signals included 250 samples each, we decided to take 1/4th of the sampling rate as a unit number. The closest power of 2 was 64. In the backward and forward directions, this means that our model included of 128 units in BiLSTM configuration.  Two fully-connented (called also dense) layers, of 32 and 1 node, followed BiLSTM layer. Between those layers, dropout layer was set as the regularization method. Dropout rate was set to 0.5. Activation function in output Fully Connected layer was Sigmoid. The selection of power of two as the unit number in the LSTM layer was supported by connecting CNNs and LSTM in the next step. The selection of 32 nodes in the first Fully Connected layer was supported by trials with different sizes of 16, 32, 64 and 128. That number in that BiLSTM configuration gave the best results.

The learning rate was set to 0.001, the optimizer was Adam and the loss function was binary cross-entropy. 

\begin{figure}
    \centering
    \includegraphics[width=1\linewidth]{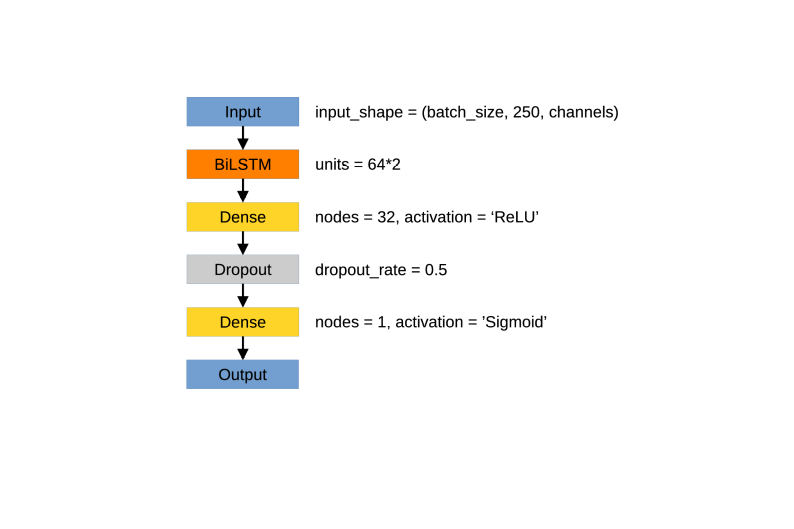}
    \caption{LSTM detailed model architecture with parameters of specific layers}
    \label{fig:lstm_model_architecture}
\end{figure}

Using the COGN-26 data set, the model had 50,753 parameters. After training using the FULL-256 data set, the model had 168,513 parameters. Detailed architecture is given in Fig.~\ref{fig:lstm_model_architecture}.

\subsection{1D-CNN}
\begin{figure}
    \centering
    \includegraphics[width=1\linewidth]{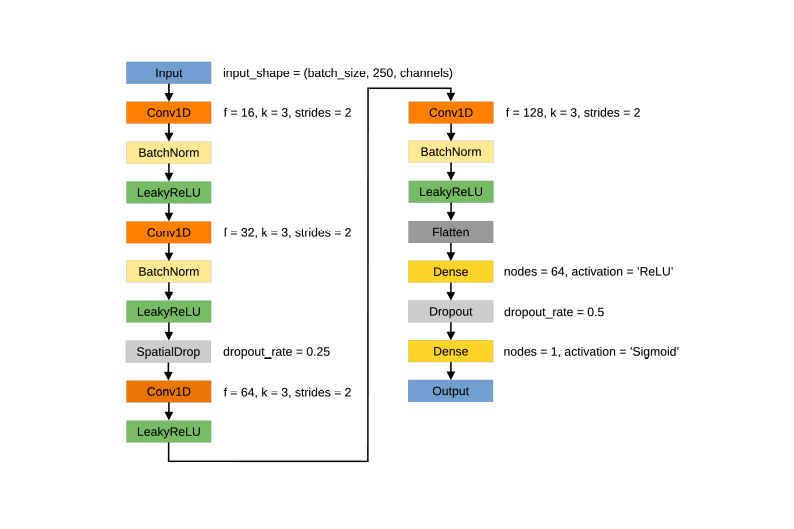}
    \caption{Detailed 1D-CNN model architecture with parameters of specific layers: f - number of filters, k - kernel size, pool sizes and dropout rates}
    \label{fig:model}
\end{figure}

The proposed CNN model included of 4 convolutional layers. The layer is the main element of Convolutional Neural Network. It contains a set of filters which adjust their parameters during the model training phase. The LeakyReLU activation layer was used after each convolutional layer to provide non-linearity\cite{schmidhuber2015deep}. Moreover, the Batch Normalization layer was used in each block of convolution containing a convolution layer and an activation layer. The purpose of Batch Normalization is to normalize data in batch to enhance learning speed and performance. Batch Normalization was neglected in the third block of convolution because Spatial Dropout (called SpatialDrop in Fig.~\ref{fig:model}) with the dropout rate of 0.25 was used before. Spatial Dropout is a method of regularization that drops randomly features learned by convolution layer during training to reduce overfitting\cite{spatialDropout}.  Instead of using pooling layers, strided convolution was applied. It can provide simpler architecture with better accuracy in some applications\cite{springenberg2014striving}. In the case of proposed CNN model it was the best choice in terms of achieved accuracy. The Flatten layer was set in front of two Fully Connected layers,
which are responsible for binary classification of features extracted by convolutional layers.
The dropout layer was used between Fully Connected layers as regularization method. It deactivates randomly weights of certain parameters during the training process to reduce overfitting\cite{dropout}. The dropout rate was set to 0.5.

For the 1D-CNN model the loss function and optimizer remain the same as for the EEGNet and LSTM-based model. The learning rate was reduced to 0.00001 from the default value of 0.001.

The numbers of parameters in the model for COGN-26 and FULL-256 data sets were: 165,649 and 176,689 respectively. Figure \ref{fig:model} shows the model architecture in detail.

\subsection{1D-CNN-LSTM}

It was proved that 1D-CNN-LSTM can be applied to the EEG signals successfully. It was reported that this kind of approach can be beneficial for epileptic seizures classification \cite{gaowei2020cnnlstm} and motor imagery classification \cite{motorimagery2022hongli}.

\begin{figure}
    \centering
    \includegraphics[width=1\linewidth]{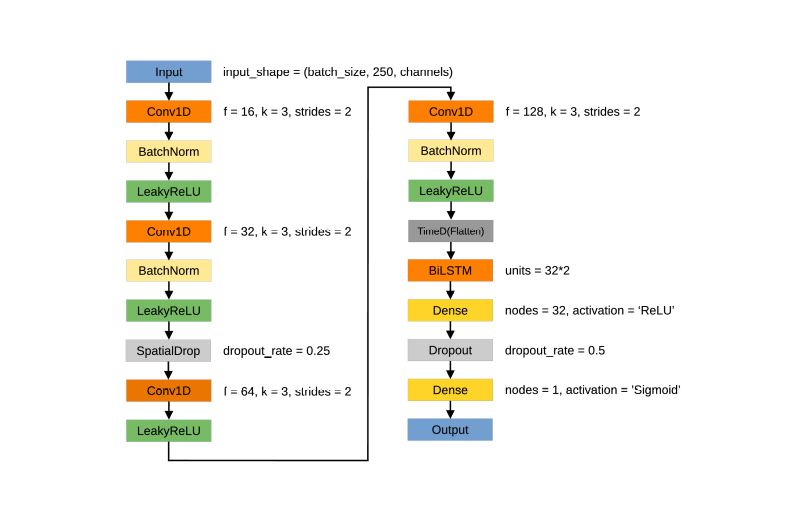}
    \caption{1D-CNN-LSTM detailed model architecture with parameters of specific layers: f - number of filters, k - kernel size, pool sizes and dropout rates.}
    \label{fig:cnn_lstm_model}
\end{figure}

A decision was made to connect 1D-CNN network model with the LSTM one described in the previous sections. In order to pass the Flatten output as input to the BiLSTM layer, and mantain the same model weights for all output data, the Time Distributed layer was used (referenced in Figure ~\ref{fig:cnn_lstm_model} as TimeD. Moreover as data are processed in CNN layers and the input size for LSTM part is already reduced, we decided to reduce number of nodes in first Fully Connected layer from 64 to 32. This resulted in model architecture shown on Figure ~\ref{fig:cnn_lstm_model}.

The numbers of parameters in the model for the COGN-26 and FULL-256 data sets were: 77,777 and 88,817 respectively. The learning rate, optimizer and loss function were set as for 1D-CNN model.

\subsection{Evaluation metrics}

Validation accuracy was selected as the main performance metric due to the fact that the balanced data sets were used for the binary classification. Validation loss was also monitored during the model designing phase. F1-score, precision and recall averaged over 6 folds are also reported for all tested models. Mentioned metrics are defined as follows\cite{evaluationMetrics}:
\begin{equation}
    Accuracy = \frac{TP+TN}{TP+TN+FP+FN}
\end{equation}

Here TP is defined as True Positives, TN - True Negatives, FP - False Positives and FN - False Negatives.
\begin{equation}
    Precision = \frac{TP}{TP+FP}
\end{equation}

Precision quantifies the accurate prediction of positive labels within the total predicted labels belonging to the positive class.

\begin{equation}
    Recall = \frac{TP}{TP+FN}
\end{equation}

Recall is a measure of the number of positive labels that are correctly classified.

\begin{equation}
    F1 = \frac{2*Precision*Recall}{Precision+Recall} = \frac{2*TP}{2*TP+FP+FN}
\end{equation}

F1 is defined as the harmonic mean between the recall and precision values.

\section*{Results}

All architectures were tested using keras and tensorflow 2.15 packages with Python 3.11. The
hardware used for testing was an Intel i7-based machine with 64GB of DDR5 RAM. The machine
was also equipped with the Nvidia GeForce RTX 4070-based graphics card with 12GB of RAM. The operating system
was Ubuntu 23.10. None of the setup elements were overclocked.

EEGNet was chosen as a reference because of the well documented architecture. The 6-fold cross validation procedure was perofmed using the model. The results for each fold and validation metrics such as: accuracy, loss, F1-score, precision and recall as well as their average values with standard deviations are presented in Tables ~\ref{tab:eegnet_metrics_full} and ~\ref{tab:eegnet_metrics_cogn} - for the FULL-256 and COGN-26 data sets.
On the average after the 6-fold cross-validation EEGNet obtained 0.7615 and 0.7646 accuracy respectively. In terms of precision as well as recall and F1-score with all average metrics exceeding 0.75 on both data sets model can be considered as a good reference point.

\begin{table}[ht]
\caption{EEGNet Validation Metrics for FULL-256 data set for each}
\centering
\begin{tabular}{llllll}
\toprule
\textbf{Fold} & \textbf{ACC} & \textbf{Loss} & \textbf{F1-score} & \textbf{Precision} & \textbf{Recall} \\
\midrule
1 & 0.7917 & 0.4186 & 0.8016 & 0.7652 & 0.8417 \\
2 & 0.7958 & 0.562  & 0.8293 & 0.7126 & 0.9917 \\
3 & 0.7563 & 0.6193 & 0.7053 & 0.8917 & 0.5833 \\
4 & 0.6458 & 0.9923 & 0.7195 & 0.5956 & 0.9083 \\
5 & 0.8375 & 0.3612 & 0.8465 & 0.8022 & 0.8958 \\
6 & 0.7417 & 0.655  & 0.6575 & 0.9754 & 0.4958 \\
\midrule
\textbf{Avg} & 0.7615 & 0.6014 & 0.7600   & 0.7905 & 0.7861 \\
\textbf{Std.} & 0.0658 & 0.2230  & 0.0764 & 0.1336 & 0.1989 \\
\bottomrule
\end{tabular}
\label{tab:eegnet_metrics_full}
\end{table}

\begin{table}[ht]
\caption{EEGNet Model Validation Metrics COGN-26 data sets for each fold}
\centering
\begin{tabular}{llllll}
\hline
\textbf{Fold} & \textbf{ACC} & \textbf{Loss} & \textbf{F1} & \textbf{Precision} & \textbf{Recall} \\
\midrule
1 & 0.5312 & 0.6879 & 0.5455 & 0.5294 & 0.5625 \\
2 & 0.7729 & 0.8835 & 0.8149 & 0.6877 & 1.0000 \\
3 & 0.6812 & 0.7774 & 0.5321 & 1.0000 & 0.3625 \\
4 & 0.9042 & 0.2666 & 0.8996 & 0.9450 & 0.8583 \\
5 & 0.7917 & 0.3944 & 0.8270 & 0.7071 & 0.9958 \\
6 & 0.9063 & 0.2442 & 0.9036 & 0.9295 & 0.8792 \\
\midrule
\textbf{Avg} & 0.7646 & 0.5423 & 0.7538 & 0.7998 & 0.7764 \\
\textbf{Std.} & 0.1427 & 0.2755 & 0.1705 & 0.1856 & 0.2579 \\
\bottomrule
\end{tabular}
\label{tab:eegnet_metrics_cogn}
\end{table}
The LSTM model with only one LSTM layer followed by dropout was chosen as second reference point. The results 6-fold cross-validation and validation metrics such as: accuracy, loss, F1-score, precision and recall as well as their average values with standard deviations are presented in Tables~\ref{tab:lstm_metrics_full} and ~\ref{tab:lstm_metrics_cogn}. On the FULL-256 data set precision, recall and F1-score achieved the averaged over folds values above 0.72. The averaged ACC for this case was 0.7250 on the full set of channels. The model performed worse than EEGNet in terms of all described metrics. On the data set containing only 26 electrodes it performed the worst of all compared models with the cross-validated accuracy of 0.6833. It achievied also the worst cross-validated accuracy for both data sets.

\begin{table}[ht]
\caption{LSTM Model Validation Metrics on FULL-256 data set for each fold}
\centering
\begin{tabular}{llllll}
\toprule
\textbf{Fold} & \textbf{ACC} & \textbf{Loss} & \textbf{F1} & \textbf{Precision} & \textbf{Recall} \\
\midrule
1 & 0.6417 & 1.3984 & 0.7346 & 0.5833 & 0.9917 \\
2 & 0.7479 & 0.7055 & 0.7881 & 0.6798 & 0.9375 \\
3 & 0.7771 & 0.7453 & 0.8022 & 0.7209 & 0.9042 \\
4 & 0.5417 & 2.434  & 0.5000 & 0.5500 & 0.4583 \\
5 & 0.8854 & 0.3187 & 0.896  & 0.8200 & 0.9875 \\
6 & 0.7563 & 0.6777 & 0.6777 & 1.0000 & 0.5125 \\
\midrule
\textbf{Avg} & 0.725  & 1.0466 & 0.7331 & 0.7257 & 0.7986 \\
\textbf{Std.} & 0.1187 & 0.7644 & 0.1355 & 0.1658 & 0.2454 \\
\bottomrule
\end{tabular}
\label{tab:lstm_metrics_full}
\end{table}

\begin{table}[ht]
\caption{LSTM Model Validation Metrics on COGN-26 data set for each fold}
\centering
\begin{tabular}{llllll}
\toprule
\textbf{Fold} & \textbf{ACC} & \textbf{Loss} & \textbf{F1} & \textbf{Precision} & \textbf{Recall} \\
\midrule
1 & 0.6542 & 3.0136 & 0.7422 & 0.5916 & 0.9958 \\
2 & 0.75 & 4.276 & 0.8 & 0.6667 & 1.0000 \\
3 & 0.5729 & 5.629 & 0.5393 & 0.5854 & 0.500 \\
4 & 0.5437 & 5.277 & 0.6803 & 0.5236 & 0.9708 \\
5 & 0.8125 & 0.8726 & 0.8421 & 0.7273 & 1.0000 \\
6 & 0.7667 & 2.267 & 0.6957 & 1.0000 & 0.5333 \\
\midrule
\textbf{Avg} & 0.6833 & 3.5559 & 0.7166 & 0.6824 & 0.8333 \\
\textbf{Std.} & 0.1101 & 1.8403 & 0.1063 & 0.1709 & 0.2458 \\
\bottomrule
\end{tabular}
\label{tab:lstm_metrics_cogn}
\end{table}

The 6-fold cross validation procedure was applied for the 1D-CNN model. The results for each fold and validation metrics such as: accuracy, loss, F1-score, precision and recall as well as their average values with standard deviations are presented in Tables~\ref{tab:1d_cnn_metrics_full} and ~\ref{tab:1d_cnn_metrics_cogn}. The averaged over folds accuracy for this model using the FULL-256 data set was 0.7682 which can be considered as a result comparable to that of the EEGNet model. On cognitive electrodes subset it achieved 0.8094 accuracy which outperforms all described architectures for this case. Also in terms of F1-score, precision and recall this model performs the best in the research for the COGN-26 data set.

\begin{table}[ht]
\caption{1D-CNN Model Validation Metrics on FULL-256 data set for each fold}
\centering
\begin{tabular}{llllll}
\toprule
\textbf{Fold} & \textbf{ACC} & \textbf{Loss} & \textbf{F1} & \textbf{Precision} & \textbf{Recall} \\
\midrule
1 & 0.8758 & 0.3077 & 0.8745 & 0.8782 & 0.8708 \\
2 & 0.7917 & 0.5703 & 0.8227 & 0.7160  & 0.9667 \\
3 & 0.8375 & 0.4113 & 0.8169 & 0.9355 & 0.7250  \\
4 & 0.6042 & 1.6410  & 0.6494 & 0.5828 & 0.7333 \\
5 & 0.7625 & 0.5687 & 0.8034 & 0.6853 & 0.9708 \\
6 & 0.7375 & 0.5966 & 0.6519 & 0.9672 & 0.4917 \\
\midrule
\textbf{Avg} & 0.7682 & 0.6826 & 0.7698 & 0.7942 & 0.7931 \\
\textbf{Std.} & 0.0947 & 0.4828 & 0.0954 & 0.1547 & 0.1827 \\
\bottomrule
\end{tabular}
\label{tab:1d_cnn_metrics_full}
\end{table}

\begin{table}[ht]
\caption{1D-CNN Model Validation Metrics on COGN-26 data set for each fold}
\centering
\begin{tabular}{llllll}
\toprule
\textbf{Fold} & \textbf{ACC} & \textbf{Loss} & \textbf{F1} & \textbf{Precision} & \textbf{Recall} \\
\midrule
1 & 0.8833 & 0.2768 & 0.8848 & 0.8740 & 0.8958 \\
2 & 0.7583 & 0.9046 & 0.8041 & 0.6761 & 0.9917 \\
3 & 0.6854 & 0.7771 & 0.5519 & 0.9588 & 0.3875 \\
4 & 0.9312 & 0.2232 & 0.9281 & 0.9726 & 0.8875 \\
5 & 0.8188 & 0.4445 & 0.7981 & 0.9005 & 0.7167 \\
6 & 0.7792 & 0.5075 & 0.7166 & 1.0000 & 0.5583 \\
\midrule
\textbf{Avg} & 0.8094 & 0.5223 & 0.7806 & 0.8970 & 0.7396 \\
\textbf{Std.} & 0.0886 & 0.271 & 0.1341 & 0.1179 & 0.2312 \\
\bottomrule
\end{tabular}
\label{tab:1d_cnn_metrics_cogn}
\end{table}

The 6-fold cross validation procedure was applied for the hybrid 1D-CNN-LSTM model. The results for each fold and mentioned earlier validation metrics such as: accuracy, loss, F1-score, precision and recall as well as their average values with standard deviations are presented in Tables ~\ref{tab:1d_cnn_lstm_metrics_full} and ~\ref{tab:1d_cnn_lstm_metrics_cogn}. The averaged over folds validation accuracy for this model trained using the FULL-256 data set was 0.7726. This was the best accuracy result for the full set of channels of all approaches discussed in this paper. On the cognitive electrodes subset it achieved 0.7556 accuracy which outperforms only the plain LSTM model in this case. For the COGN-26 data set the results are worse than those of 1D-CNN and comparable with EEGNet.

\begin{table}[ht]
\centering
\caption{1D-CNN-LSTM Model Validation Metrics for FULL-256 data set for each fold}
\begin{tabular}{llllll}
\toprule
\textbf{Fold} & \textbf{ACC} & \textbf{Loss} & \textbf{F1} & \textbf{Precision} & \textbf{Recall} \\
\midrule
1 & 0.7292 & 0.6919 & 0.7789 & 0.658  & 0.9542 \\
2 & 0.7500 & 1.069  & 0.7993 & 0.6676 & 0.9958 \\
3 & 0.8583 & 0.3907 & 0.8373 & 0.9831 & 0.7292 \\
4 & 0.7167 & 0.9444 & 0.7247 & 0.7047 & 0.7458 \\
5 & 0.8875 & 0.2449 & 0.8945 & 0.8419 & 0.9541 \\
6 & 0.6938 & 1.214  & 0.5638 & 0.9794 & 0.3958 \\
\midrule
\textbf{Avg} & 0.7726 & 0.7591 & 0.7664 & 0.8058 & 0.7958 \\
\textbf{Std.} & 0.0803 & 0.3852 & 0.1144 & 0.1510 & 0.2268 \\
\bottomrule
\end{tabular}
\label{tab:1d_cnn_lstm_metrics_full}
\end{table}

\begin{table}[ht]
\caption{1D-CNN-LSTM Model Validation Metrics for COGN-26 data set for each fold}
\centering
\begin{tabular}{llllll}
\toprule
\textbf{Fold} & \textbf{ACC} & \textbf{Loss} & \textbf{F1} & \textbf{Precision} & \textbf{Recall} \\
\midrule
1 & 0.7688 & 0.5999 & 0.7861 & 0.7312 & 0.9758 \\
2 & 0.7542 & 1.3300 & 0.8027 & 0.6704 & 1.0000 \\
3 & 0.7250 & 1.2670 & 0.6207 & 1.0000 & 0.4500 \\
4 & 0.7771 & 0.5214 & 0.8165 & 0.6939 & 0.9917 \\
5 & 0.7271 & 0.566 & 0.7207 & 0.7380 & 0.7042 \\
6 & 0.7813 & 0.7457 & 0.7200 & 1.0000 & 0.5625 \\
\midrule
\textbf{Avg} & 0.7556 & 0.8383 & 0.7445 & 0.8056 & 0.7807 \\
\textbf{Std.} & 0.0247 & 0.3648 & 0.0732 & 0.1526 & 0.2423 \\
\bottomrule
\end{tabular}
\label{tab:1d_cnn_lstm_metrics_cogn}
\end{table}

The averaged 6-fold cross-validated metrics for all models are reported in Tab.~\ref{tab:metrics_full} for the FULL-256 data set and in Tab.~\ref{tab:metrics_cogn} for the COGN-26 data set. It can be seen that the worst model for classification from the full set of electrodes is the one-layer LSTM-based model. The other models obtained comparable results in terms of accuracy, while the best one was the 1D-CNN-LSTM hybrid model. On the other hand for the signal collected from subset of cognitive electrodes in terms of validation metrics of accuracy, loss, F1-score and precision the 1D-CNN-based model outperformed all other approaches with the accuracy of 0.8094, the F1-score value of 0.7806 and the precision close to 0.8970.  

\begin{table}[ht]
\caption{Evaluation of Metrics for Different Models for the FULL-256 data set. The best result for every metric is reported in bold.}
\centering
\begin{tabular}{llllll}
\toprule
\textbf{Model} & \textbf{ACC} & \textbf{Loss} & \textbf{F1} & \textbf{Precision} & \textbf{Recall} \\
\midrule
EEGNet & 0.7615 & 0.6014 & 0.75995 & 0.79045 & 0.7861 \\
LSTM & 0.7250 & 1.0466 & 0.7331 & 0.7257 & \textbf{0.7986} \\
1D-CNN & 0.7682 & \textbf{0.6826} & \textbf{0.7698} & 0.7942 & 0.7931 \\
1D-CNN-LSTM & \textbf{0.7726} & 0.7592 & 0.7664 & \textbf{0.8058} & 0.7958 \\
\bottomrule
\end{tabular}
\label{tab:metrics_full}
\end{table}

\begin{table}[ht]
\caption{Evaluation of Metrics for Different Models using the COGN-26 data set. The best result for each metric is reported in bold.}
\centering
\begin{tabular}{llllll}
\toprule
\textbf{Model} & \textbf{ACC} & \textbf{Loss} & \textbf{F1} & \textbf{Precision} & \textbf{Recall} \\
\midrule
EEGNet & 0.7646 & 0.5423 & 0.7538 & 0.7998 & 0.7764 \\
LSTM & 0.6833 & 3.5559 & 0.7166 & 0.6824 & \textbf{0.8333} \\
1D-CNN & \textbf{0.8094} & \textbf{0.5223} & \textbf{0.7806} & \textbf{0.8970} & 0.7396 \\
1D-CNN-LSTM & 0.7556 & 0.8383 & 0.7445 & 0.8056 & 0.7807 \\
\bottomrule
\end{tabular}
\label{tab:metrics_cogn}
\end{table}

\section*{Discussion}

There are known approaches of using convolutional neural networks in biometrics \cite{prakash2022baed} and other cybernetical tasks \cite{daoui2023aucfsr}, more and more of them in the EEG signal classification \cite{prakash2022baed}. More and more often deep learning methods are applied in the biomedical engineering systems to help patients with numerous of disorders like sleep apneua \cite{kandukuri2023constant}

The aim of this paper was to compare the effectiveness of four different architectures in the EEG signal classification originating from a psychological experiment involving Guided Imagery. There were used the EEGNet, LSTM, 1D-CNN and 1D-CNN-LSTM approaches in the case of dense array amplifier setup using 256 electrodes and the so-called cognitive setup using 26 electrodes. 

Training all of these models is relatively fast, does not require extensive
resources, and as a result can be incorporated into less demanding computational
environments after training using different data. What is also beneficial is that in spite of the
fact that the EEG signal can vary in time and between subjects, it is possible to train the
model with great accuracy using smaller segments of 1 second instead of 1 minute or even
the full-length signal. Benefit of this work is also that all the models make use of all 256 EEG channels to learn features and its simplified version of 26 cognitive channels. 

Indeed, the results obtained in this study show that the manual feature extraction (EEG bands, wavelets
etc.) can be neglected while using the CNN-based, LSTMs and hybrid models architectures. 

Simple filtration and interpolation of the signal seem to be sufficient. The binary signal classifiers described above perform well on
raw data, resulting in the level of accuracy comparable to that of state-of-the art methods
and to our previous paper on Generalized Linear Model in EEG signal classification
\cite{zemla2023-glm}. 

In case of the full signal collection recorded from 256 electrodes the 1D-CNN-LSTM performs best in terms of accuracy and precision. Almost as good as the one above is 1D-CNN, especially that it has better loss and F1-score values. One layer LSTM accuracy is the worst in this experiment, however still higher than 0.70 with the best recall of 0.79. 
The reference model EEGNet has the accuracy of 0.76 (compared to the best discussed here 0.77) and generally lower characteristics in the case of remaining three metrics. The collection and comparison of all results of the discussed classifiers are presented in Tab.~\ref{tab:metrics_full}.

In the case of the signal collected from 26 cognitive electrodes evidently the best one is the proposed 1D-CNN model achieving 80\% accuracy with the best loss, F1 and precision characteristics. The one-layer LSTM has much lower accuracy (68\%) but its recall is the highest reaching 0.83. The accuracy of the EEGNet reached 0.76 and 1D-CNN-LSTM 0.75 which were lower by 5\%  compared with the best one 1D-CNN. The other parameters like F1 and precision are of the same order of value, relatively similar but none is as good as that for 1D-CNN-LSTM. The collection and comparison of all results of the discussed classifiers are presented in Tab.~\ref{tab:metrics_cogn}.

Better performance on 26 electrodes (accuracy of 81\% for 1D-CNN vs 77\% for 1D-CNN-LTSM) can be the result of putting more influential data for feature extraction and automatically selecting those of greater significance for the task than the manually selected subset of 256 electrodes or in special case all of them.

Thus it was proved that from the computational point of view it is even more beneficial to collect fewer data for such tasks and expanding the cap to 256 electrodes does not always add a significant value. 

There is still place for improving those models by training them with more data from more subjects. There is also need to test if best models work well for the data gathered from female subjects. Also finding new architecture for this task can be a way to reducing number of parameters of the model. It needs to be investigated how other electrodes subsets, like 10-20 international system\cite{chatrian1985ten} can affect performance of classification using the described architecture.

Another aspect of improvements that can be applied is the parameter tuning for the models. In our opinion, based on the previous experience \cite{wojcik2023investigating} this could increase the accuracy of the models by 3\%-5\%. 

Then, there can be designed more complex hybrid architectures, involving other methods of EEG signal analysis \cite{kawala2014innovative, kahankova2017non} or eg. the fuzzy logic approach \cite{mikolajewska2017computational, prokopowicz2017fuzzy} 

The research presented here can shed new light on the engineering of new brain-computer interfaces with application for psycho-therapists and neuro-therapists using the relaxation techniques and Guided Imagery method. 

\section*{Acknowledgements}

The authors would like to thank Ewa Lewandowska and Marek Rudziński of VRMed for
fruitful discussions, inspiration and possibility of working together on new ideas.

\section*{Author contributions statement}

F.P.: Principal Investigator of the project, research project conceptualization at applied neuroinformatics dimension, concept of convolutional neural networks application, data processing, data analysis, manuscript preparation, models design and implementation, EEG recordings;
G.M.W.: Key Investigator of the research, research idea and methodological support, manuscript preparation;
K.W.: EEG recordings, work in the laboratory;
A.K.: statistical consultation;
G.S.: research project conceptualization at psychological dimension;
K.Z.: research project conceptualization at psychological dimension and implementing Guided Imagery relaxation
technique.

\section*{Data availability statement}
The raw data supporting the conclusions of this manuscript will be made available by the
authors, without undue reservation, to any qualified researcher. To obtain the data please contact Filip Postepski using e-mail address: filip.postepski@mail.umcs.pl.

\section*{Additional information}
\textbf{Funding:} This study was financed by Maria Curie-Sklodowska University as beneficiary of "UMCS Mini-Grants" programme in the academic year 2022/23.\\
\textbf{Conflict of interest:} No conflicts of interest are declared.\\
\textbf{Ethics approval and consent to participate:} The studies involving human participants were reviewed and approved by the Maria Curie-
Skłodowska University Bioethical Commission. The participants provided their written informed consent to participate in this study.

\bibliographystyle{unsrtnat}
\bibliography{references}  

\end{document}